\documentclass[10pt,twocolumn,letterpaper]{article}

\usepackage{iccv}              

\usepackage[dvipsnames]{xcolor}
\usepackage{subcaption}

\definecolor{iccvblue}{rgb}{0.21,0.49,0.74}
\usepackage[pagebackref,breaklinks,colorlinks,allcolors=iccvblue]{hyperref}

\usepackage{multirow}
\usepackage{comment} 
\newcommand{\PAR}[1]{\vspace{-0.2eM}\vskip4pt \noindent{\bf #1}}

\def\paperID{10874} 
\def\confName{ICCV}
\def\confYear{2025}

\title{\vspace*{-0.5em}Self-Supervised Sparse Sensor Fusion for Long Range Perception}
\author{
\begin{tabular}[t]{c}
Edoardo Palladin$^{*1}$ \qquad
Samuel Brucker$^{*1}$ \\[0.1em]
Filippo Ghilotti$^1$ \qquad
Praveen Narayanan$^1$ \qquad
Mario Bijelic$^{1,2}$ \qquad
Felix Heide$^{1,2}$ 
\end{tabular}\\[0.2em]
\begin{tabular}[t]{c}
\small{$^1$Torc Robotics} \qquad
\small{$^2$Princeton University} \qquad
\small{$^*$Equal contribution} \\[0.1em]
\small{\url{https://light.princeton.edu/LRS4Fusion}}
\end{tabular}
}

\begin{document}
\twocolumn[{
\maketitle

\vspace{-2.0em}
}]

\begin{abstract}
Outside of urban hubs, autonomous cars and trucks have to master driving on intercity highways. 
Safe, long-distance highway travel at speeds exceeding 100 km/h demands perception distances of at least 250 m, which is about five times the 50–100m typically addressed in city driving, to allow sufficient planning and braking margins. Increasing the perception ranges also allows to extend autonomy from light two-ton passenger vehicles to large-scale forty-ton trucks, which need a longer planning horizon due to their high inertia.
However, most existing perception approaches focus on shorter ranges and rely on Bird’s Eye View (BEV) representations, which incur quadratic increases in memory and compute costs as distance grows. To overcome this limitation, we built on top of a sparse representation and introduced an efficient 3D encoding of multi-modal and temporal features, along with a novel self-supervised pre-training scheme that enables large-scale learning from unlabeled camera-LiDAR data. Our approach extends perception distances to 250 meters and achieves an $\emph{26.6\%}$ improvement in mAP in object detection and a decrease of $\emph{30.5\%}$ in Chamfer Distance in LiDAR forecasting compared to existing methods, reaching distances up to 250 meters.
\vspace{-1.0em}
\end{abstract}

\vspace{-1.5em}
\begin{figure}[t!]
     \centering
     \includegraphics[width=\columnwidth]{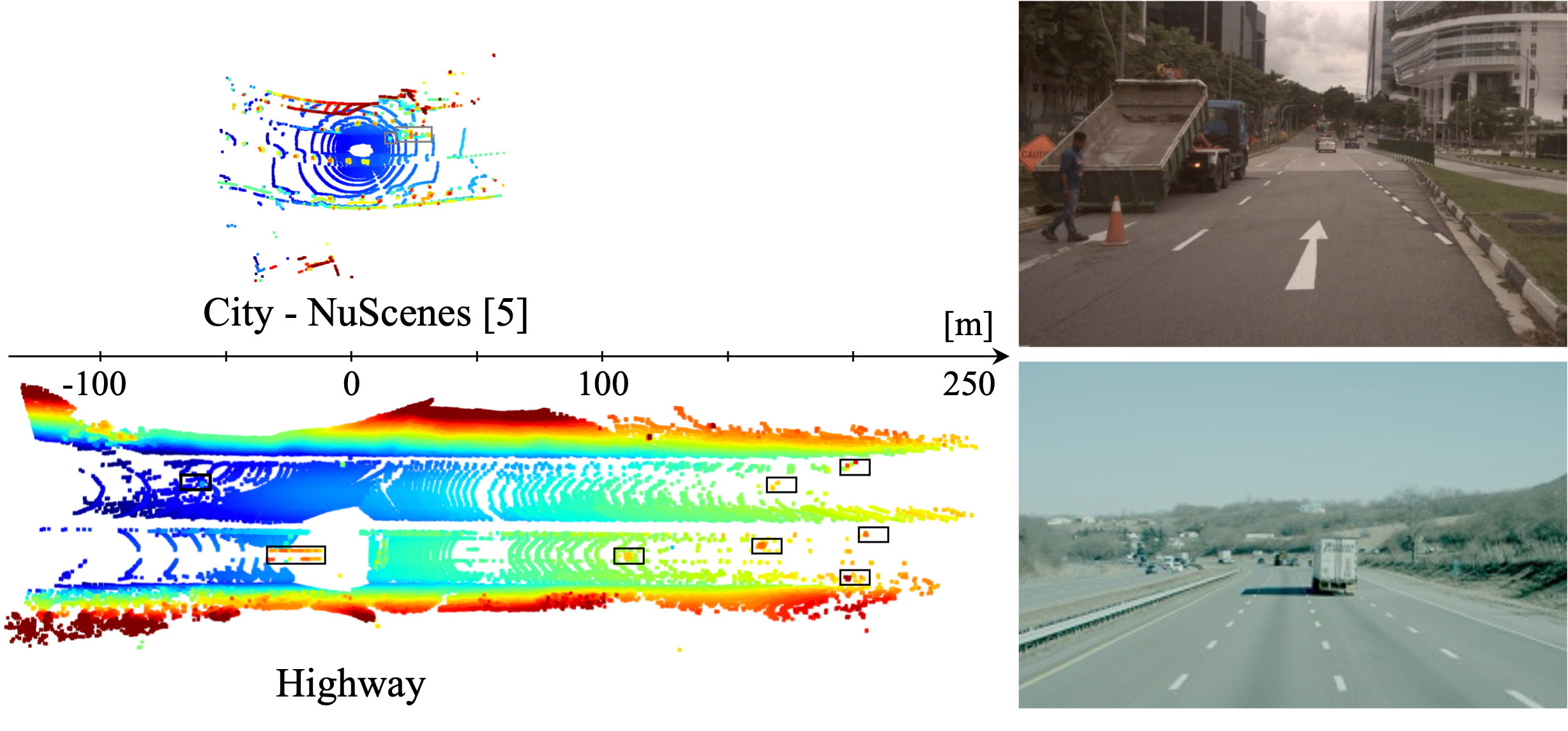}
     \caption{Autonomous vehicles, especially trucks with long braking distances, require long planning horizons for efficient and safe driving in highway scenarios (bottom). This requires extending the typical perception range from $50$–$100m$ (top~\cite{caesar2020nuscenesmultimodaldatasetautonomous}) to beyond $250m$ where dense representations struggle. We introduce a sparse voxel fusion approach for efficient and accurate 3D scene understanding enabling processing of long-range LiDAR-camera data that leverages spatio-temporal context up to 250m. The proposed model outputs \emph{depth, occupancy, velocity, future LiDAR forecast, accurate object detection}, see Fig. \ref{fig:OD_qual} and Fig. \ref{fig:QualitativeNuscenes}.}
     \label{fig:Teaser}\vspace{-1eM}
 \end{figure}
\section{Introduction}
Autonomous vehicles rely on precise perception of their surroundings for scene understanding, prediction, and planning. Today's most successful methods that allow for 360$^\circ$ perception rely on BEV features to perform 3D object detection \cite{samfusion, yang2022deepinteraction}, semantic occupancy prediction \cite{tong2023scene}, tracking \cite{zhang2022mutr3d} and planning \cite{hu2023planning, weng2024drive}. The methods operate on BEV features that encode critical information about the 3D environment, derived from single or multiple sensors that perceive the surroundings.
However, most existing BEV-based methods focus on short-term planning with ranges below 50m \cite{liu2023bevfusion,li2022bevformer}, making them well-suited for robo-taxi applications in urban environments but insufficient for long-term planning needs, particularly necessary for highway driving and in the context of robo-trucking, where long braking distances demands the perception of objects and planning decisions at long-ranges, to ensure safe, strategic planning beyond 70~m, which existing approaches do not address \cite{hu2023planning, weng2024drive, jiang2023vad} (Fig. \ref{fig:Teaser}). 

Extending BEV features to cover longer ranges presents significant challenges, as computational complexity and memory footprint grow rapidly. A dense BEV feature map that holds all the information alone grows quadratically in memory with detection ranges. Relying on computations to projecting camera information into a unified representation, such as Lift-Splat-Shoot \cite{philion2020lift}, also increases computational complexity quadratically with range.
Therefore, in this work we propose a method that enables BEV features far beyond surround LiDAR and camera data by using a sparse voxel implementation, addressing the limitations of current approaches in long-range perception. 

\begin{figure}[t!]
\vspace{-0.5em}
    \centering
    \includegraphics[width=\columnwidth]{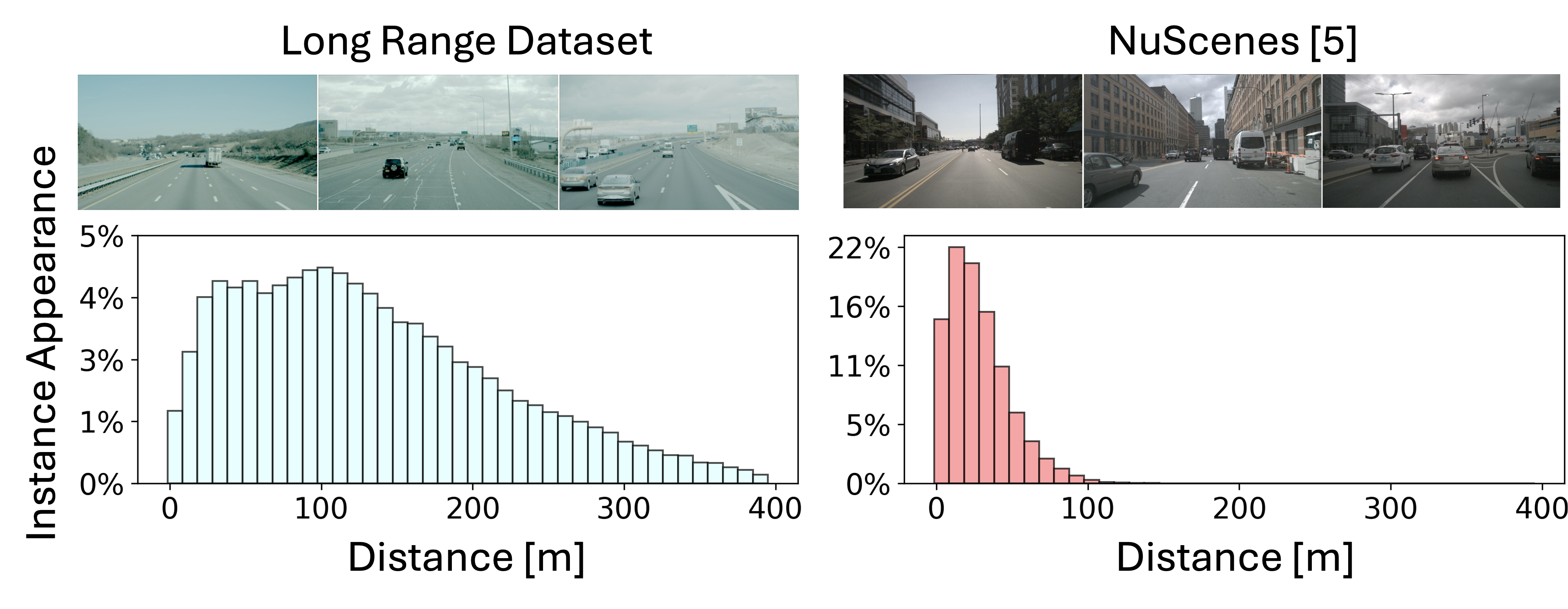}\vspace{-1.0em}
    \caption{
Experimental Long-Range Dataset. To assess the proposed method, we capture and annotate a long-range dataset, showing the instance distribution in the training split alongside NuScenes \cite{caesar2020nuscenesmultimodaldatasetautonomous}. While only NuScenes is shown, other popular datasets like Argoverse2 \cite{argoverse} and ONCE \cite{once} follow similar urban distributions, with a concentrated peak of instances in the very close range, unlike our long-range setting with boxes up to 400m.
}
    \label{fig:NumIstances}\vspace{-0.5cm}
\end{figure}

To prevent the training data corpus from growing equally, as objects become increasingly sparse at extended distances, we propose a self-supervision approach. As shown in Fig.~\ref{fig:NumIstances}, the frequency of object instances decreases significantly with distance, so simply increasing the spatial coverage forces a steep increase in required labeled data, which is both costly and time-consuming. Recent approaches have embraced self-supervised pre-training strategies \cite{vidar, uno}, no longer requiring large labeled ground truth datasets. By encoding both current and past sensor data, these methods leverage temporal information to predict future states of the environment. This forecasting is supervised by reconstructing sensor inputs and letting the model learn robust encodings from the natural evolution of the scene over time.

However, such existing self-supervised pre-training approaches are limited to a single modality (e.g., camera-only \cite{vidar} or LiDAR-only \cite{uno}), which restricts their ability to generalize in multi-sensor systems, most commonly deployed in autonomous vehicles. Fusing multiple modalities, such as surround LiDAR and cameras, requires learning to identify complementary information and sparse 3D data from LiDAR with dense, high-resolution imagery from cameras, which we solve with a sparse local attention scheme.
The proposed approach enables occupancy prediction, depth prediction, lidar forecasting and object detection.

To train the proposed method for these diverse tasks, we devise a new self-supervised pre-training method that enables long-range multimodal perception without relying on labeled data. Through direct supervision of future LiDAR point clouds and velocities, we demonstrate that the pre-training leads to high performance on tasks such as LiDAR forecasting and 3D object detection for ranges up to $250m$.

\noindent We make the following contributions:
\begin{itemize}
    \item We introduce a long-range LiDAR-camera fusion approach built with a computationally efficient fully sparse voxel representation.
    \item We devise a self-supervised training approach that incorporates temporal information from past frame history and with self-supervised pretraining providing spatio-temporal context without labels.
    \item We validate that the method achieves state-of-the-art performance on the LiDAR forecasting task, improving Chamfer Distance by up to 30.5\% for ranges up to 250m range and by up to 29\% on the NuScenes Dataset for up to 51.2m, on the hardest $3 sec$ future horizon prediction.
    \item We validate the method for Object Detection,  improving by $\emph{26.6\%}$ ($\emph{+11.06 mAP}$) on long detection ranges of up to $\emph{250m}$ over existing methods.
\end{itemize}

\section{Related Work}

\PAR{3D Scene Representations} 
are the underpinning of efficient 3D environment perception.
Existing work has modeled 3D space as voxels, where each voxel is characterized by an assigned vector \cite{zhou2018voxelnet}.
While voxel-based representations excel at capturing detailed 3D structures for tasks such as 3D semantic occupancy prediction, including LiDAR segmentation \cite{cao2022monoscene, wang2023openoccupancy, wei2023surroundocc}, they lack computational efficiency due to the large number of voxels involved. However, since the vertical dimension typically carries less critical information than the horizontal dimensions, BEV-based approaches streamline the representation by encoding height data within each grid cell \cite{lang2019pointpillars}.
BEV-based approaches are popular in 3D object detection \cite{li2022bevformer, samfusion, liu2023bevfusion}, but also semantic occupancy prediction \cite{huang2024selfocc}, trajectory prediction \cite{gu2023vip3d, teeti2022vision} and planning \cite{hu2023planning, jia2023driveadapter, jia2023think, weng2024drive}. 
As a compromise between voxel and BEV representations, the TPV (Tri-Perspective View) representation has been proposed \cite{huang2023tri} which relies on three perpendicular cross-planes to represent the 3D scene, initializes query sets on these planes to gather features from images, and exchanges features across views using attention. 
Other works explore sparse BEVs for 3D detection \cite{liu2023sparsebev} and sparse voxels \cite{li2024fully}, drawing on efficient implementations from sparse pointcloud processing for tasks like occupancy prediction \cite{tang2024sparseocc} and other downstream applications \cite{wang2024distillnerf}.
GaussianFormer \cite{huang2024gaussianformer} propose a latent 3D gaussian representation that is splatted to voxels for 3D Occupancy prediction. 

\PAR{Multi-Modal Perception} 
aim to enrich LiDAR feature maps by integrating semantic information from camera images~\cite{vora2020pointpainting,wang2021pointaugmenting,yin2021multimodal}. These methods were foundational in combining data from different sensor types. Subsequent research has explored cross-modal feature-level fusion, further refining this integration by directly merging features from both modalities~\cite{yoo20203d,xu2021fusionpainting}.
To address the challenge of accurately projecting RGB camera features into the LiDAR space \cite{li2022unifying} leverage deformable attention mechanisms~\cite{zhu2020deformable} to create a unified 3D voxel representation, blending both modalities within a shared spatial framework.
More recent approaches operating within the BEV space have enabled fusion of features from multiple sensors in a common reference frame (typically the LiDAR BEV perspective). The aggregated features are then processed by task-specific decoders for applications such as 3D object detection~\cite{liang2022bevfusion, cai2023bevfusion4d, li2022bevformer, liu2023petrv2, yin2024fusion}, lane estimation~\cite{HDMapnet, BEVSegformer, liu2023petrv2}, object tracking~\cite{hu2023planning}, semantic segmentation~\cite{liu2023bevfusion, li2022bevformer, liu2023petrv2}, and planning~\cite{hu2023planning}. This multi-task and multi-modal setup benefits from additional supervision and regularization, improving overall performance across various perception tasks.
However, current BEV-based methods still face limitations in projecting detailed camera features into BEV coordinates, primarily due to their reliance on monocular depth estimation techniques~\cite{ku2018defense} or Lift-Splat-Shoot (LSS) methods~\cite{liu2023bevfusion}, which estimate depth for camera features and may introduce inaccuracies.

\PAR{Depth Estimation} is a core capability of camera-only geometric perception. Approaches \cite{liu2023bevfusion,li2022bevformer} utilize variants of Lift-Splat-Shoot (LSS) \cite{philion2020lift} to lift 2D camera features into a 3D space. Alternatively, some methods \cite{samfusion} focus on using predicted dense maps and subsequently projecting image features into the 3D LiDAR frame based on the estimated depth. Such depth maps could be predicted by applying widely used monocular depth estimation \cite{yin2023metric3d, piccinelli2024unidepth, bochkovskii2024depth, yang2024depth}, though with limited accuracy, especially at long ranges due to the inherent scale ambiguity. Stereo depth estimation \cite{raftstereo, walz2023gated, brucker2024cross, xu2023iterative}, in contrast, has shown to be more accurate, particularly over long distances, but requires overlapping camera setups for effectiveness.
In configurations that include both camera and LiDAR sensors, depth completion methods \cite{zhang2023completionformer, ognidc, park2020non, tang2020learning} can be employed to achieve the highest-precision depth predictions. These methods project sparse LiDAR points into image space and utilize image features to interpolate and complete the sparse depth information effectively. We build on this idea to predict dense depth for accurate image feature projection, but devise a lightweight architecture that extends ranges beyond 250m.

\PAR{Large Scale Self-Supervised Pretraining}
methods have employed contrastive approaches \cite{chen2020improved, he2020momentum, khosla2020supervised, tian2020contrastive} and masked signal modeling \cite{devlin2018bert, he2022masked, wei2022masked, xie2022simmim}. However, method that rely on pre-training for autonomous driving, which demands semantic understanding, 3D structure, and temporal modeling, have only recently been explored.
VoxelMAE \cite{hess2023masked} extends Masked AutoEncoders to LiDAR data for object detection. UniPAD \cite{yang2024unipad} builds on this by reconstructing color and depth from masked multi-modal inputs. ALSO \cite{boulch2023also} uses surface reconstruction from present-time LiDAR rays as a pre-training task. ViDAR \cite{vidar} explores pre-training with temporal modeling by reconstructing future LiDAR from current and past images. Similarly, UnO \cite{uno} learns a 4D spatiotemporal occupancy field for the reconstruction of future LiDAR from past LiDAR. 
Recently, DistillNeRF \cite{wang2024distillnerf} has demonstrated the distillation of foundation models such as DINOv2 \cite{oquab2023dinov2} as pretraining for enhancing semantic understanding of scenes.
In contrast to existing work, we learn multi-model perception systems and encode 3D structure, semantic understanding, and temporal modelling, focusing on long ranges. 

\section{LRS4Fusion}
\begin{figure*}[t!]
    \centering
    \includegraphics[width=2.05\columnwidth]{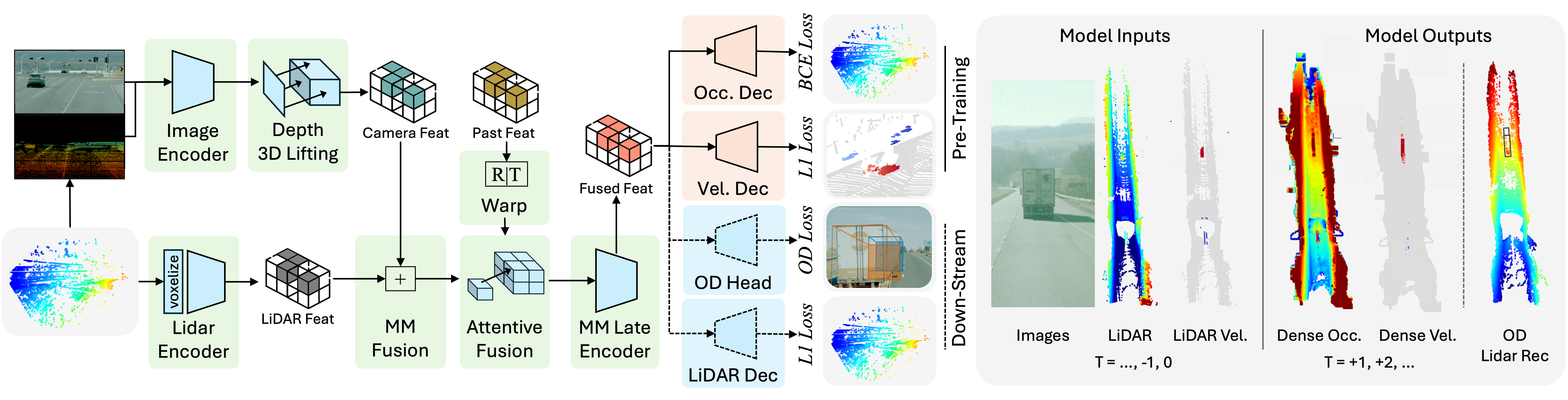}
    \vspace{-0.5em}
    \caption{LRS4Fusion: Sparse Multi-Modal Self-Supervision and Fusion. Camera features are lifted to 3D through accurate depth-maps and joined with LiDAR features in a unified sparse representation. The resulting features are fused through a custom sparse attention with past features before being further processed by the Multi-Modal Late Encoder. During the pre-training stage, final features are passed to the custom sparse occupancy decoder and velocity decoder. In contrast to existing SOTA methods that focus on pre-training mono-modality backbones, our approach aims at pre-training multimodal encoders. The method produces depth, occupancy, velocity, and LiDAR at future frames, along with object detection (OD) predictions.}
    \label{fig:Overview}\vspace{-0.7em}
\end{figure*}

In this section, we introduce the proposed \textbf{L}ong-\textbf{R}ange \textbf{S}elf-\textbf{S}upervised \textbf{S}parse \textbf{S}ensor \textbf{Fusion} (LRS4Fusion) approach, which is illustrated in Fig. \ref{fig:Overview}. To support prediction distances of up to 250 meters, we rely on a sparse voxel representation that allows to exploit multi-modal and temporal cues. We train the method with a novel self-supervised training scheme. We first introduce the multi-modal feature extraction and fusion using sparse voxels in Sec. \ref{sec:feat_extraction}, then the self-supervised pre-training scheme in Sec. \ref{sec:supervision}.

\subsection{Sparse Sensor Fusion}
\label{sec:feat_extraction}

\PAR{The Camera Encoder} extracts features from each multi-view image and projected lidar pointcloud. Therefore, we
project the lidar pointcloud $P$ onto every camera frame \( I^{\text{RGB}}_i \) for each camera \( i \), producing a corresponding sparse depth image \( I^{D}_i \). This results in a 4-channel image defined as
$I^{\text{RGBD}}_i = \left[ I^{\text{RGB}}_i, I^{D}_i \right]$,
where \( I^{\text{RGB}}_i \in \mathbb{R}^{H \times W \times 3} \) and \( I^{D}_i \in \mathbb{R}^{H \times W \times 1} \), yielding \( I^{\text{RGBD}}_i \in \mathbb{R}^{H \times W \times 4} \). Image features $F$ are extracted using $f_{\text{img}}$ a multi-scale feature pyramid based on Vim \cite{visionmamba} as follows,
\begin{equation}
F_1^i, F_2^i, F_3^i, F_4^i = f_{\text{img}}(I^{RGBD}_i).
\end{equation}
We additionally concatenate a camera embedding, derived from intrinsic and extrinsic calibration matrices, to each flattened patch sequence—small non-overlapping image regions encoded by the Vim \cite{visionmamba} backbone—to explicitly incorporate camera-specific geometry into feature extraction. Encoded features are extracted at 4 different Vim depths and then decoded to generate feature maps at multiple scales. The resulting outputs are fed into an FPN network \cite{FPN_OD} to generate features that are later lifted to 3D.
\PAR{The Depth Estimation} build on top of the 
multi-scale features $F = \{F_1^i, F_2^i, F_3^i, F_4^i\} $, which are passed to the depth model $f_{depth}$ to predict dense depth $D_i$ for each frame $i$. The model employs a multi-scale recurrent architecture that iteratively refines depth predictions by integrating sparse LiDAR depth and image features with increasing resolution. For each scale, a small backbone $\mathcal{B}$ extracts context $h_t$ and confidence $C_{\text{inp}}$ features which guide the refinement process, that is \vspace*{-8pt}
\begin{equation}
    C_{\text{inp}}, h_t = \mathcal{B}(F)
\end{equation}
We employ a convolutional update block that uses a Minimal Gated Unit (MGU) \cite{zhou2016minimal} to improve the depth map. The MGU, designed with a single forget gate for simplicity, updates depth gradients by adjusting the hidden state based on current depth estimates, whereas more widely applied GRU-based networks \cite{raftflow, raftstereo, ognidc} typically rely on separate update and reset gates to control state adjustments. This consolidation of gates reduces computational load and parameter count by one third, enhancing efficiency while maintaining effective control for accurate depth predictions.
Each iteration refines the depth map by first estimating a depth gradient $\nabla d_t = F_g(h_t)$ from context features, in a depth gradient network $F_g$, then merging that gradient with the previous depth estimate in a depth integration module. This module balances sparse LiDAR depth with image-derived gradients, producing a refined map that incorporates both sensor measurements and visual context. Formally, the updated depth is given by:
\begin{equation}
d_{t+1} = d_t - \Delta d,
\end{equation}
where the correction term $\Delta d$ is computed as
\begin{equation}
\Delta d = f_{\text{update}}\left( \nabla d_t - g, \, (d_t - s_d) \odot M, \, C_{dg}, \, C_{\text{inp}} \right).
\end{equation}
Here, $\nabla d_t$ is the gradient of the current depth estimate $d_t$, $g$ is the predicted depth gradient, $s_d$ is the sparse depth measurement, $M$ is the valid sparse mask, $C_{dg}$ is the confidence in the depth gradient, and $f_{\text{update}}$ represents the convolutional operations within the integration module.

By integrating depth and gradient discrepancies with confidence information across multiple scales, the network incrementally refines the depth map over successive passes, initially ensuring global consistency and later incorporating high-frequency details while smoothing out inaccuracies. The confidence information is predicted by $C_{dg} = f_{\text{conf}}(h_t)$, where $f_{\text{conf}}$ is a learned confidence head applied to the hidden state $h_t$ in the MGU update block. 

\PAR{Lifting 2D Features into 3D} is performed as first step by projecting each image frame using the camera matrix $K$ and predicted depth $D_i$ to 3D points $\mathbf{X}_C$ from pixel coordinates $(u,v)$, as $\mathbf{X}_C = D_i(u,v)K^-1(u,v,1)$. The predicted coordinates are casted into sparse voxels and features are aggregated per voxel cell, that results in a sparse representation in the form of $F_C^i = [\mathbf{F}_C, \mathbf{X}_C]^i$ for $i = 1,...,N$ and $\mathbf{F}_C \in \mathbb{R}^{N,F}, \mathbf{X}_C \in \mathbb{N}^{N,3}$, where $N$ is the number of hidden features and $F$ the hidden feature dimension.

\PAR{The LiDAR Encoder} processes the LiDAR scan $P$, which is first voxelized and encoded using $f_{\text{lid}}$. We implement $f_{\text{lid}}$ as voxel-wise PointNet \cite{qi2017pointnet}. Features are then extracted through sparse convolutions, followed by a sparse U-Net \cite{shi2020pointsparts3dobject}.
Similar to the camera branch, this process yields a sparse representation $F_L = [\mathbf{F}_L, \mathbf{X}_L]$, where $\mathbf{F}_L \in \mathbb{R}^{M,F}$ and $\mathbf{X}_L \in \mathbb{N}^{M,3}$, with $M$ being the number of occupied voxels and $F$ the feature dimension.

\PAR{Camera-LiDAR Fusion} combines the sparse voxels from both modalities in a unified sparse voxel space by concatenating the features from both modalities in the Sparse Fusion Module $f_{MM}$. Features are first fed into batch norm layers to normalize values from different encoding branches. During the concatenation, for voxels that are empty in either of the two modalities, zeros are appended. After the concatenation the features are passed into a sparse convolution module as a first fusion step. The resulting fused features form a single sparse representation $F_{LC} = [\mathbf{F}_{LC}, \mathbf{X}_{LC}]$ and $\mathbf{F}_{LC} \in \mathbb{R}^{Q,F}, \mathbf{X}_{LC} \in \mathbb{N}^{Q,3}$ where $Q = M + N - O$ is the number of occupied voxels, $O$ is the number of overlapping Camera and LiDAR features and $F$ the hidden feature dimension.

\PAR{Late Sparse Encoding} is applied as subsequent step, where the resulting features $F_{LC}$ are processed through a sequence of Completion Blocks and Contextual Aggregation Blocks $f_{\text{con}}$, following \cite{tang2024sparseocc}. A pyramid of features is then assembled, where the final multi-scale features are fused within the sparse representation. Compared to previous work \cite{tang2024sparseocc}, the \emph{proposed representation remains sparse along all scales}, further reducing the memory footprint and enabling the use of finer-grained discretization. The resulting latent embedding consists of sparse voxels at 4 different scales $V = [V_1, V_2, V_3, V_4]$. Multiple scales captures coarse and fine-grained information separately, preserving details while maintaining global context. This enables specialized processing before strategic fusion, while larger voxels help fill gaps and propagate information across sparse areas for better long-range perception.

\PAR{Temporal Sparse Fusion} is applied within the Late Sparse Encoding to the second smallest $V_2$ Voxel representation only. 
Therefore we utilize the current $t_0$ and last $t_{-1}$ timestamp. For abbreviation we drop the $^2$ exponent and write $V^{t_0}$ and $V^{t_{-1}}$.
To align the voxel maps we need both the rigid body transformation $R\mid T^{t_{-1}\rightarrow t_{1}}$ between the timestamps and the velocity per voxel to correct the vehicle movements. The velocity are observed directly from the FMCW LiDAR measurements and we accumulate the velocity per voxel $v_q^{t_{-1}}$. The $R\mid T^{t_{-1}\rightarrow t_{1}}$ are obtained from the vehicle odometry. 
Past features $V^{t_{-1}} = [\mathbf{V}^{t_{-1}}, \mathbf{X^{t_{-1}}}]$ are warped to $V^{t_0'} = [\mathbf{V}^{t_{-1}}, \mathbf{X}^{t_0'}]$, where the new positions are computed as

\begin{equation}
    \mathbf{X}_q^{t_0'} = (\mathbf{X}_q^{t_{-1}} + \mathbf{v}_q^{t_{-1}} dt) \mathbf{R|T}^{t_{-1} \rightarrow t_0}. 
\end{equation}

Here, $q = 1, \dots, Q$ represents the number of past occupied voxels cells.

To maintain sparsity transformed features can not be concatenated with the current features as it we would sequentially add more and more occupied voxels to the latent representation over time, completely neglecting the memory efficiency of the sparse representation.

\begin{figure}[t!]
    \centering
    \includegraphics[width=0.65\columnwidth]{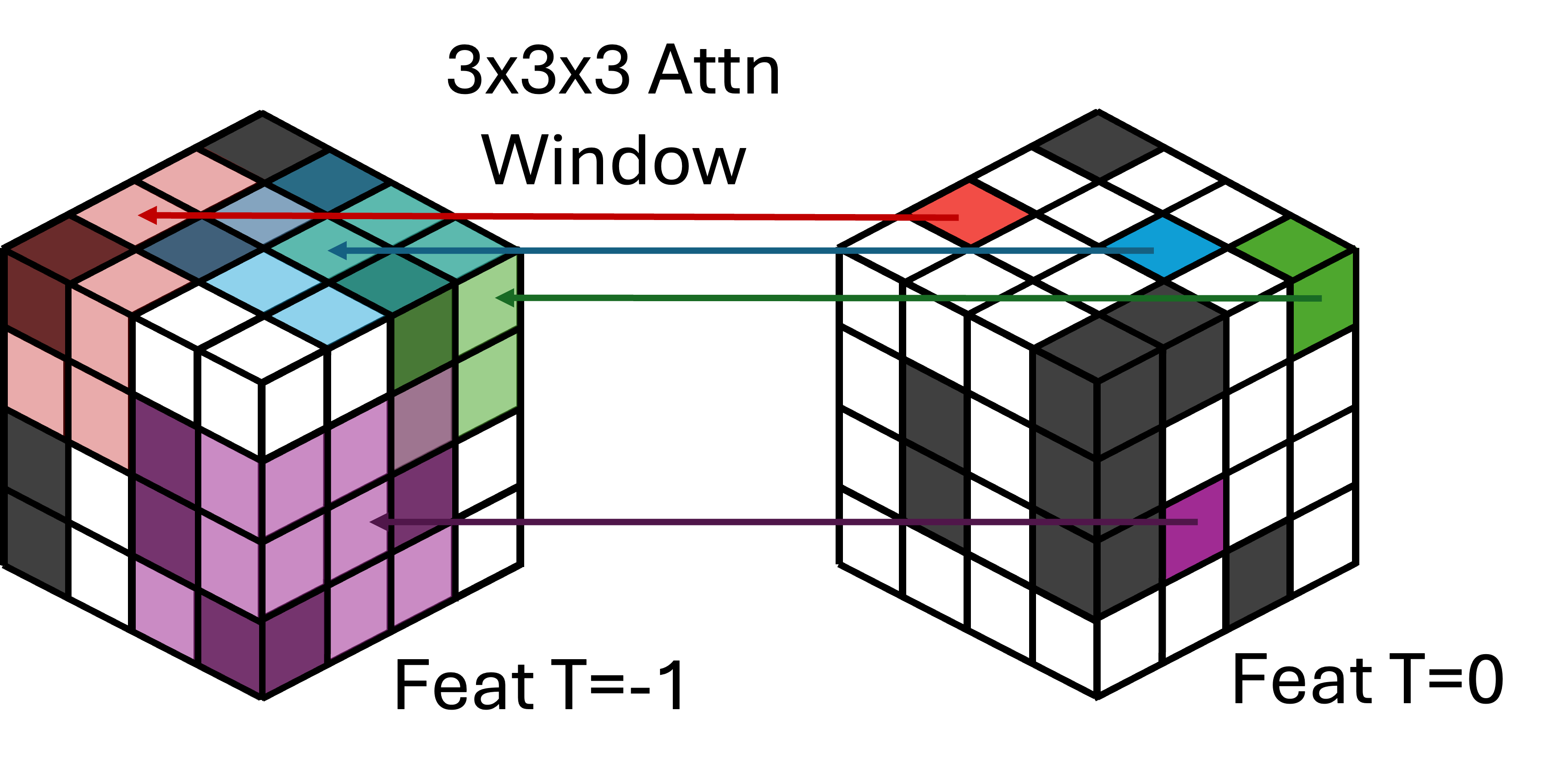}
    \vspace{-0.5em}
    \caption{Sparse Window Attention. Occupied voxels in the current frame $T=0$ attends to a window of occupied voxels in the previous timestamp $T=-1$. The attention is computed between each query in the current frame and all occupied key,values inside the attention window centered at the same location but at the previous timestamp. The method ensure that the final number of occupied voxels does not explode when the number of occupied previous voxels is high. In the example, we consider an attention window of 3x3x3, \textcolor{red}{Red} Query attends to 3 voxels in the past frame, \textcolor{blue}{Blue} to 3, \textcolor{green}{Green} to 2 and \textcolor{purple}{Purple} to 4.}
    \label{fig:TemporalWarping}\vspace{-0.7em}
\end{figure}

Therefore, we introduce a novel sparse windowed attention layer, see Fig.~\ref{fig:TemporalWarping}. Inspired by local attention mechanism, each occupied voxel at the current timestamp $t_0$ attends to a 3D window of voxels in the previous warped timestamp $t_{-1}$. The operation directly operates in the sparse representation, without the need of converting the features to a dense BEV grid or dense voxel volume. This approach significantly reduces computational cost while enabling the windowed attention to capture information beyond local neighborhoods, effectively capturing moving actors and correcting misalignments between past voxels transformed into the current reference.
Hence, sparse history enhanced features are calculated as
\begin{equation}
 V_* = \sum_{V^{t_0'}\in J_s} \text{softmax}\biggl(\frac{V^{t_0}\,(V^{t_0'})^{T}}{\sqrt{d}}\biggr)V^{t_0'},
\end{equation}
where $Js$ is the set of neighboring voxels inside the attention sampling window and $d$ the softmax normalization factor representing the dimensionality of the hidden features. Note, basing the mechanism on top of the occupied voxels $V^{t_0}$ the queries don't include unoccupied cells ensuring that the amount of occupied voxels does not explode.

\subsection{Self-Supervision}
\label{sec:supervision}
Due to the quadratically shrinking resolution for long ranges and the diminished number of objects at very long distances (Fig. \ref{fig:NumIstances}), training requires an extensive amount of data. To tackle this challenge, we self-supervise the embedding by a loss composed of a reconstruction loss for sparse occupancy and sparse velocity.

\begin{figure}[t!]
    \centering
    \includegraphics[width=0.9\columnwidth]{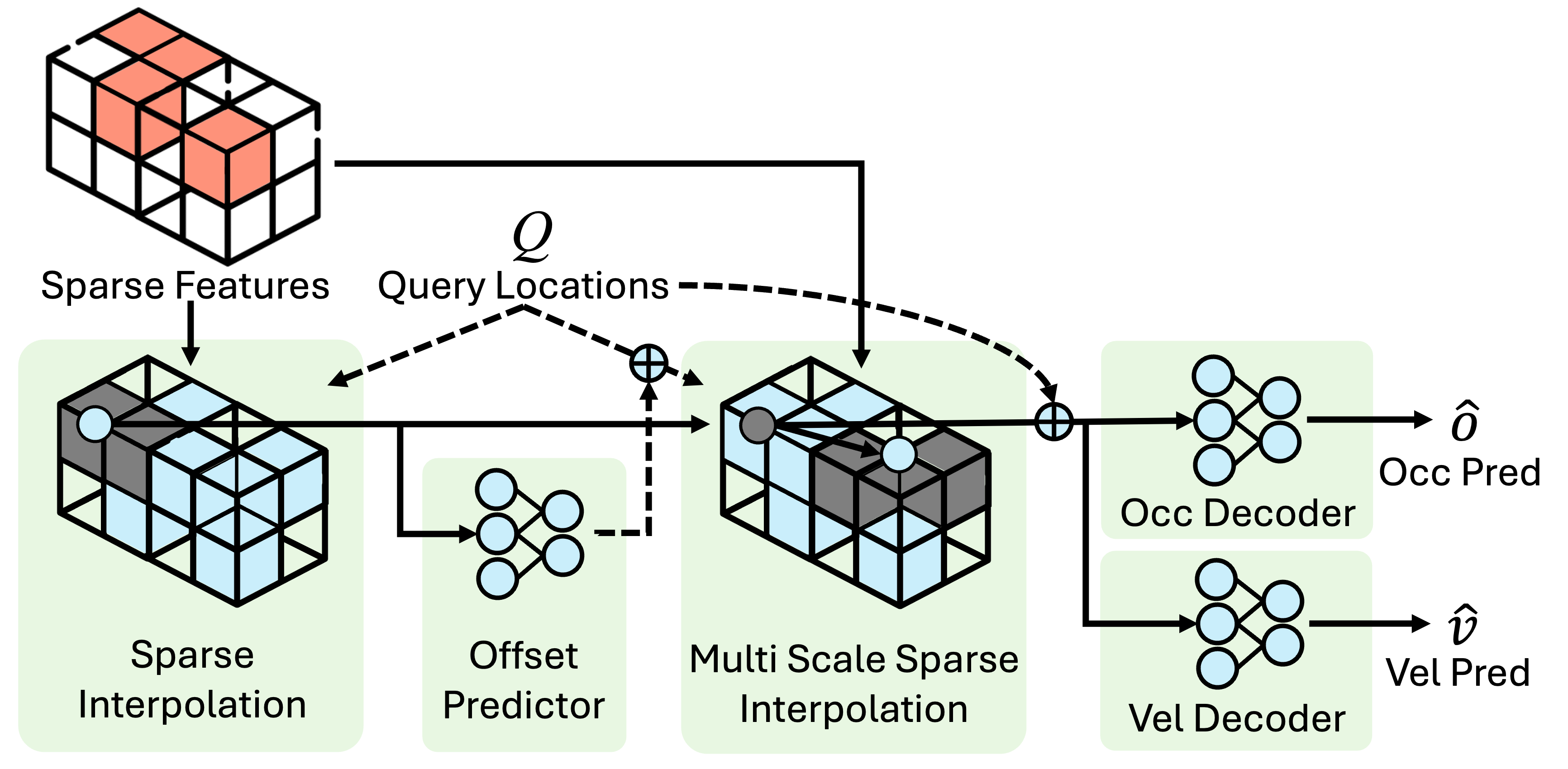}
     \caption{Occupancy and Velocity Decoder. The proposed decoder takes a 4D query point, interpolates nearby occupied voxels to refine the interpolation position based on temporal information and decodes the features from the first and second sampling to generate occupancy and velocity predictions.}
    \label{fig:DensityDecoder}
    \vspace{-0.5cm}
\end{figure}

\PAR{Sparse Occupancy and Velocity Decoder} \label{sec:density_decoder} are auxiliary prediction heads predicting a dense geometric and dynamic representation from the voxel embeddings and are shown Fig. \ref{fig:DensityDecoder}. Both can be trained at scale from self supervision alone from arbitrary driving recordings without ground truth labels. 
Both take as input a four-dimensional query point $Q$ ($x, y, z, t$), where $x, y, z$ are spatial coordinates and $t$ is time, and its outputs are both density and velocity for each queried point. If the query point lies in the past or future, the current voxel representation may not accurately reflect the scene. Rather than transforming the entire voxel space, we only transform the queried point \cite{uno}, building on top of an existing rigid transformation between the current encoding and both past and future timestamp. This allows us to account for the movements of the ego-vehicle and dynamic actors. 
A lightweight neural network $f_{\text{pose}}$ predicts the new position of the query point, whether in the past or future, by using the query ($x, y, z, t$) and tri-linearly interpolated voxels at the $(x, y, z)$ location. We interpolate the $N$ nearest neighbors at the query point intersection with the voxel grid, creating a new voxel that captures the precise sub-voxel location. The new query position is calculated as $(x', y', z') = (x, y, z) + f_{\text{pose}}(V^*(x, y, z), Q)$. At the new position, voxel features are again interpolated with $N$ nearest neighbors at all latent scales. The two sets of interpolated voxel features, from both the current and new positions, are stacked and used to predict occupancy $\hat{o}$ and velocity $\hat{v}$ through two lightweight heads $f_{\text{occ}}$ and $f_{\text{vel}}$, as

\begin{equation}
\begin{aligned}
 \hat{o} = f_{\text{occ}}(V^*(x, y, z), V^*(x', y', z'), Q), \\
 \hat{v} = f_{\text{vel}}(V^*(x, y, z), V^*(x', y', z'), Q) .
 \end{aligned}
\end{equation}

GT values for self-supervision can be estimated from the recorded LiDAR scan. Occupied labels and non-zero velocities can be directly inferred by the presence and measured velocity of a LiDAR point, while all positions along one LiDAR ray are taken as unoccupied labels as the LiDAR ray travels through free space. More details on occupancy and velocity decoder are provided in the supplemental material.

\subsection{Training}
Our training strategy is divided into three stages. In the first stage, we train the image feature encoder and depth prediction modules using a combination of image reconstruction, depth supervision, and feature distillation losses. The second stage involves training the complete model with supervision for past, current, and future frames, covering occupancy and velocity reconstruction. Lastly, we train the object detection on top. Additional training details are provided in the Supplemental Material. 

\section{Dataset}

In order to overcome the range limitations of existing LiDAR-based datasets \cite{8953693, caesar2020nuscenesmultimodaldatasetautonomous, geiger2012we, sun2020scalabilityperceptionautonomousdriving} with sensors restricted to $80$ meters, we capture a new multi-modal dataset specifically tailored to heavy-duty trucking scenarios, shown in Fig. \ref{fig:NumIstances}. We equip a semi-truck with 5 synchronized OnSemi AR0820 cameras, each featuring a $1/2$-inch CMOS sensors that captures raw RCCB data at a resolution of 3848 × 2168 pixels, and arrange them to record a near 360° view at 5 Hz. The system is complemented by Aeva Aeries 4D LiDARs, capturing 3D point clouds up to $400m$ as well as radial velocity measurements at $10Hz$, synchronized with the camera feeds. The dataset includes recordings from diverse locations in Texas, New Mexico and Virginia, spanning highway and urban environments.
The captures feature a variety of natural lighting conditions and include $60,000$ unlabeled frames and $36,000$ manually annotated frames for object detection. From these, we curate $25,000$ at $5Hz$ for our training split, capturing seven object classes with the following distribution: “Passenger-Car” ($194,807$ instances, $38.21\%$), “Vehicle” ($105,116$, $20.62\%$), “SemiTruck-Trailer” ($70,495$, $13.83\%$), “Road-Obstruction” ($95,862$, $18.80\%$), “SemiTruck-Cab” ($40,982$, $8.04\%$), “Person” ($2,448$, $0.48\%$), and “Bike” ($157$, $0.03\%$). Further information is in the Supplemental Material. 

\section{Experiments}
In this section, we validate the proposed method. Specifically, we assess the quality of the depth predictions in \cref{sec:depth}, object detection in \cref{sec:object}, and LiDAR forecasting tasks in \cref{sec:forcasting}. Additionally, we report ablation studies validating the design choices of the proposed method. Further details on the experimental setup can be found in the Supplementary Material.

\begin{table}[!tb]
    \centering
	\caption{
    The proposed depth model brings boost in speed and reduction in memory footprint while achieving the same accuracy to SOTA methods. Evaluation on accumulated LiDAR point cloud from 0 to 250m, more details to the depth evaluation are provided in the supplemental material. 
    } 
    \vspace{-0.5em}
    \resizebox{\linewidth}{!}{
    \begin{tabular}{l|ccccc}
    \toprule
        \textbf{Method} &\textbf{MAE} $\downarrow$ & \textbf{RMSE} $\downarrow$ & \textbf{Runtime [ms]} $\downarrow$ & \textbf{MEM [GB]} $\downarrow$\\
        \midrule
        {Completion Former} \cite{zhang2023completionformer} & 4.98 & 12.36 & 188 & 2.1\\
        {OGNI-DC} \cite{ognidc} & 4.76 & 13.16 & 364 & 2.4 \\
        \textbf{LRS4Fusion} & \textbf{3.46} & \textbf{9.21} & \textbf{64} & \textbf{1.3}\\
    \bottomrule
    	\end{tabular}
    \label{tab:depth_model}}
\end{table} 

\subsection{Depth Evaluation}\label{sec:depth}
We first evaluate depth prediction of the proposed method and recent existing methods \cite{zhang2023completionformer} and \cite{ognidc} compared to accumulated LiDAR ground truth. The experimental setup is detailed in the Supplemental Material. The role of the depth network is to reduce inference time and reduce memory footprint without compromising accuracy. Tab. \ref{tab:depth_model} reports how the proposed architecture achieves the lowest MAE and MSE among the three methods. The proposed architecture improves MAE by 27\% and MSE by 25\%, it achieves the lowest inference time of $0.064$s and 1.3GB of memory.

\subsection{Object Detection}\label{sec:object}

\begin{table}[!tb]
    \centering
	\caption{
    3D Object Detection performances on Long Range Dataset. \textcolor{red}{$^\ddagger$} is with ViDAR \cite{vidar} pre-train.
    } 
    \vspace{-0.5em}
    \resizebox{0.9\linewidth}{!}{
    \begin{tabular}{l|c|cc}
    \toprule
        \textbf{Method} & \textbf{Modality} & \textbf{mAP} $\uparrow$ & \textbf{NDS} $\uparrow$ \\
        \midrule
        {PointPillars} \cite{lang2019pointpillars} & L & 39.31 & 41.52 \\
        {BEVFormer} \cite{li2022bevformer} & C & 23.67 & 37.99\\
        {BEVFormer} \cite{li2022bevformer} (w/ Pre-train\textcolor{red}{$^\ddagger$})  & C & 24.51 & 38.93\\
        {BEVFusion} \cite{liu2023bevfusion} & L + C & 40.10 & 48.43\\ 
        {SAMFusion} \cite{samfusion} & L + C & 41.55  & 52.44\\
        \hline
        \textbf{LRS4Fusion} (w/o Pre-train)  & L + C & 49.58 & \textbf{59.12}\\
        \textbf{LRS4Fusion} & L + C & \textbf{52.61} & 58.06\\
    \bottomrule
    	\end{tabular}
    \label{tab:OD}}
\vspace{-1.0em}
\end{table}

\begin{figure*}[t!]
     \centering
     \includegraphics[width=1.0\linewidth]{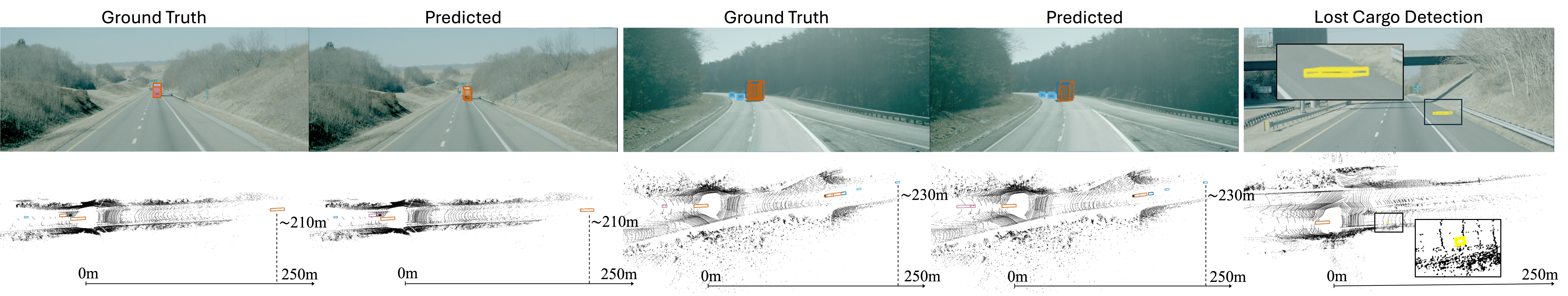}\vspace{-0.5em}
      \caption{Object Detection. The method detects vehicles and small lost cargo objects at long distances beyond 100m. 3D bounding boxes of Car, Truck-Cab, Truck-Trailer, Road Obstruction with \textcolor{cyan}{cyan}, \textcolor{blue}{blue}, \textcolor{orange}{orange}, \textcolor{yellow}{yellow} color, respectively. Please zoom in for details.}\vspace{-1em}
     \label{fig:OD_qual}
 \end{figure*}

We compare the proposed method on the OD task against recent existing single-modality and multi-modality methods on the common Mean Average Precision (mAP) and NuScenes Detection (ND) Score metrics. We project sparse features at the smallest scale into a BEV grid and pass them to a CenterPoint \cite{yin2021centerbased3dobjectdetection} head. The camera-only method BEVFormer \cite{li2022bevformer} ($23.67 mAP$) fails to achieve the performance of LiDAR models. Pre-training the feature extractor on ViDAR \cite{vidar} increases the detection accuracy by $3.55\%$ ($24.51 mAP$), reinforcing the importance of the pre-training step in the long-range scenarios. The LiDAR-only method PointPillars achieves $39.31mAP$. The fusion method BEVFusion \cite{liu2023bevfusion} performs only marginally better ($+2.01\%$) than the LiDAR-only approach, $40.10 mAP$, confirming the drawbacks of the LSS approach. SAMFusion \cite{samfusion} methods, based on depth-based 3d lifting of camera features, improves by $5.70\%$ over the LiDAR only baseline.
The proposed method ($52.61$) achieves state-of-the-art quality in the long-range dataset, improving the mAP by $26.6\%$ over the second-best method, SAMFusion, on the long-range dataset. 
Finally, we report that the proposed occupancy-velocity self-supervision improves performance by $+6.11\%$ over the proposed model without self-supervision ($49.58$). Fig. \ref{fig:OD_qual} reports qualitative detections at long ranges for vehicles and road debris.

\subsection{LiDAR Forecasting}\label{sec:forcasting}

\begin{table}[!tb]
	\caption{
    LiDAR Forecasting on Long Range Dataset. We consider a ROI of $[+100m$, $-100m]$ on the Y axis of the Ego Vehicle and $[+250m$, $-100m]$ on the X axis. 
    } 
    \vspace{-0.5em}
    \resizebox{\linewidth}{!}{
    \begin{tabular}{c|l|c|cc|cc}
    \toprule
        \textbf{History} & \multirow{ 2}{*}{\textbf{Method}} & \multirow{ 2}{*}{\textbf{Modality}} & \multicolumn{ 2}{|c}{\textbf{1s}} & \multicolumn{ 2}{c}{\textbf{3s}} \\
        \textbf{Horizon} &&& CD $\downarrow$ & L1 (m)$\downarrow$ & CD $\downarrow$ & L1 (m)$\downarrow$\\
        \hline
        \multirow{ 3}{*}{1s} & {4DOcc} \cite{khurana2023pointcloudforecastingproxy} & L & 18.933 & 4.685 & - & -\\
        & {ViDAR} \cite{vidar} & C & 58.228 & 20.209 & 51.720 & 20.688\\
        & \textbf{LRS4Fusion} & L + C & \textbf{15.821} & \textbf{3.313} & \textbf{39.031} & \textbf{3.825} \\
        \hline
        \multirow{ 3}{*}{3s} & {4DOcc} \cite{khurana2023pointcloudforecastingproxy} & L & 23.580 & 2.996 & 47.81 & 4.293\\
        & {ViDAR} \cite{vidar} & C & 57.284 & 20.141 & 56.200 & 20.531\\
        & \textbf{LRS4Fusion} & L + C & \textbf{16.382} & \textbf{2.493} & \textbf{42.932} & \textbf{4.051} \\
    \bottomrule
    	\end{tabular}
    \label{tab:lidar_forecasting}}
\vspace{-0.5em}
\end{table} 

\begin{table}[!tb]
	\caption{
    LiDAR Forecasting on NuScenes. ROI of $51.2m$ on all sides around the Ego Vehicle. \textcolor{red}{$^\dagger$} denotes results as reported in \cite{vidar}.
    } 
    \vspace{-0.5em}
    \resizebox{\linewidth}{!}{
    \begin{tabular}{c|l|c|ccccccc}
    \toprule
        History & \multirow{ 2}{*}{\textbf{Method}} & \multirow{ 2}{*}{Modality} & \multicolumn{ 6}{|c}{\textbf{Chamfer Distance} $\downarrow$} \\
        Horizon&&&  0s & 0.5s & 1s & 1.5s & 2.0s & 2.5s & 3s\\
        \midrule
        
        \multirow{ 2}{*}{0s} & {HERMES} \cite{zhou2025hermesunifiedselfdrivingworld} & C & 0.59 & - & 0.78 & - & \textbf{0.95} & - & \textbf{1.17} \\
        & \textbf{LRS4Fusion} & L + C & \textbf{0.087} & \textbf{0.348} & \textbf{0.566} & \textbf{0.748} & {0.963} & {1.205} & {1.510}\\
        \midrule
        
        \multirow{ 3}{*}{1s} & {4DOcc}\textcolor{red}{$^\dagger$} \cite{khurana2023pointcloudforecastingproxy} & L & - & 1.26 & 1.88 & - & - & - & -\\
        & {ViDAR} \cite{vidar} & C & - & 1.11 & 1.25 & 1.40 & 1.57 & 1.76 & 1.97\\
        & \textbf{LRS4Fusion} & L + C & \textbf{0.06} & \textbf{0.31} & \textbf{0.48} & \textbf{0.64} & \textbf{0.79} & \textbf{0.99} & \textbf{1.25}\\
        \midrule
        
        \multirow{ 3}{*}{3s} & {4DOcc}\textcolor{red}{$^\dagger$} \cite{khurana2023pointcloudforecastingproxy} & L & - & 0.91 & 1.13 & 1.30 & 1.53 & 1.72 & 2.11\\
        & {ViDAR} \cite{vidar} & C & - & 1.01 & 1.12 & 1.25 & 1.38 & 1.54 & 1.73\\
        & \textbf{LRS4Fusion} & L + C & \textbf{0.11} & \textbf{0.33} & \textbf{0.47} & \textbf{0.61} & \textbf{0.77} & \textbf{0.97}& \textbf{1.23}\\
    \bottomrule
    	\end{tabular}
    \label{tab:lidar_forecasting_nuscenes}}
\vspace{-0.5em}
\end{table} 

\begin{table}[!tb]
    \centering
    \caption{LiDAR Forecasting on NuScenes. \cite{uno} evaluation setting.}
    \vspace{-0.5em}
    \resizebox{\linewidth}{!}{
    \begin{tabular}{c|ccccc|c}
    \toprule
        \textbf{Method} & {SPFNet} \cite{weng2020invertingposeforecastingpipeline} & {S2Net} \cite{s2net} & {RayTracing} \cite{khurana2023pointcloudforecastingproxy} & {4D-OCC} \cite{khurana2023pointcloudforecastingproxy} & {UnO} \cite{uno} & \textbf{LRS4Fusion} \\
        \midrule        
        \textbf{NFCD} $\downarrow$ & 2.50 & 2.06 &  1.66 &1.40 & 0.89 &  \textbf{0.72}\\
        \textbf{CD} $\downarrow$ & 4.14 & 3.47 &  3.59 &  4.31 & 1.80 & \textbf{0.88} \\
        \bottomrule
    	\end{tabular}
    \label{tab:lidar_forecasting_nuscenes_uno}}
    \vspace{-1.5em}
\end{table}

We analyze the performance of our second training step by evaluating the LiDAR reconstruction accuracy at $1s$ and $3s$ into the future. Following \cite{vidar, khurana2023pointcloudforecastingproxy}, we report Chamfer distance (CD) between predicted and ground truth point cloud at $1s$, $3s$ into the future and with $0s$, $1s$, and $3s$ preceding historical horizon. We compare two LiDAR forecasting methods \cite{khurana2023pointcloudforecastingproxy, zhou2025hermesunifiedselfdrivingworld} and the camera method \cite{vidar}.

\PAR{NuScenes Evaluation}
We evaluates the LiDAR forecasting task on the NuScenes dataset \cite{caesar2020nuscenesmultimodaldatasetautonomous} following the more comprehensive protocol of \cite{vidar} in Tab. \ref{tab:lidar_forecasting_nuscenes} and following \cite{uno}, condensing the performance into a single metric, in \ref{tab:lidar_forecasting_nuscenes_uno}.
The proposed method achieves state-of-the-art results, with a Chamfer distance of $0.48$ ($+61.6\%$ over the second best model) on the $1s$ in, $1s$ out tasks, $1.25$ ($+36.5\%$) on the $1s$ in $3s$ out tasks, $0.47$ ($+58.0\%$) on the $3s$ in $1s$ out tasks and $1.23$ ($+28.9\%$) on the $3s$ in $3s$ out tasks. We also compare the performance of the proposed method on the no-history ($0s$ history horizon) task with the recent Hermes \cite{zhou2025hermesunifiedselfdrivingworld} work based on large Vision Language Model (VLM): we achieve state-of-the-art results - $0.566$ on $1s$ forecasting - with future horizon less then $2s$, where the complex reasoning capabilities of VLMs are able to better forecast, but also compute-heavy.

\PAR{Long-range Evaluation}
Tab. \ref{tab:lidar_forecasting} evaluates the method on the proposed long-range dataset. 
4D-Occ struggle to extract temporal information from the history horizon due to the large motion between frame in the highway scenarios; this is visible by the drop in performance between $1s$, CD $16.87$, and $3s$, CD $23.58$, history horizon.
ViDAR \cite{vidar} is less affected by large scene movements due to its use of expensive deformable attention and multi-frame reasoning but still struggles with accurate long-range 3D geometry, as monocular surround cameras lack the necessary depth cues and geometric constraints for estimating distant structures. This is reflected by the large CD, $56.2$ for $3s$ input $3s$ output, $58.23$ for $1sec$ in $1sec$ out, in the full range setting. 
Instead, the proposed method is able to exploit the multi-modal input and to effectively extract temporal clues from the history horizon, achieving CD of 15.821 on $1sec$ in $1sec$ out and 42.932 on $3sec$ in $3sec$ out.

\begin{figure}[t!]
    \centering
    \includegraphics[width=1.0\columnwidth]{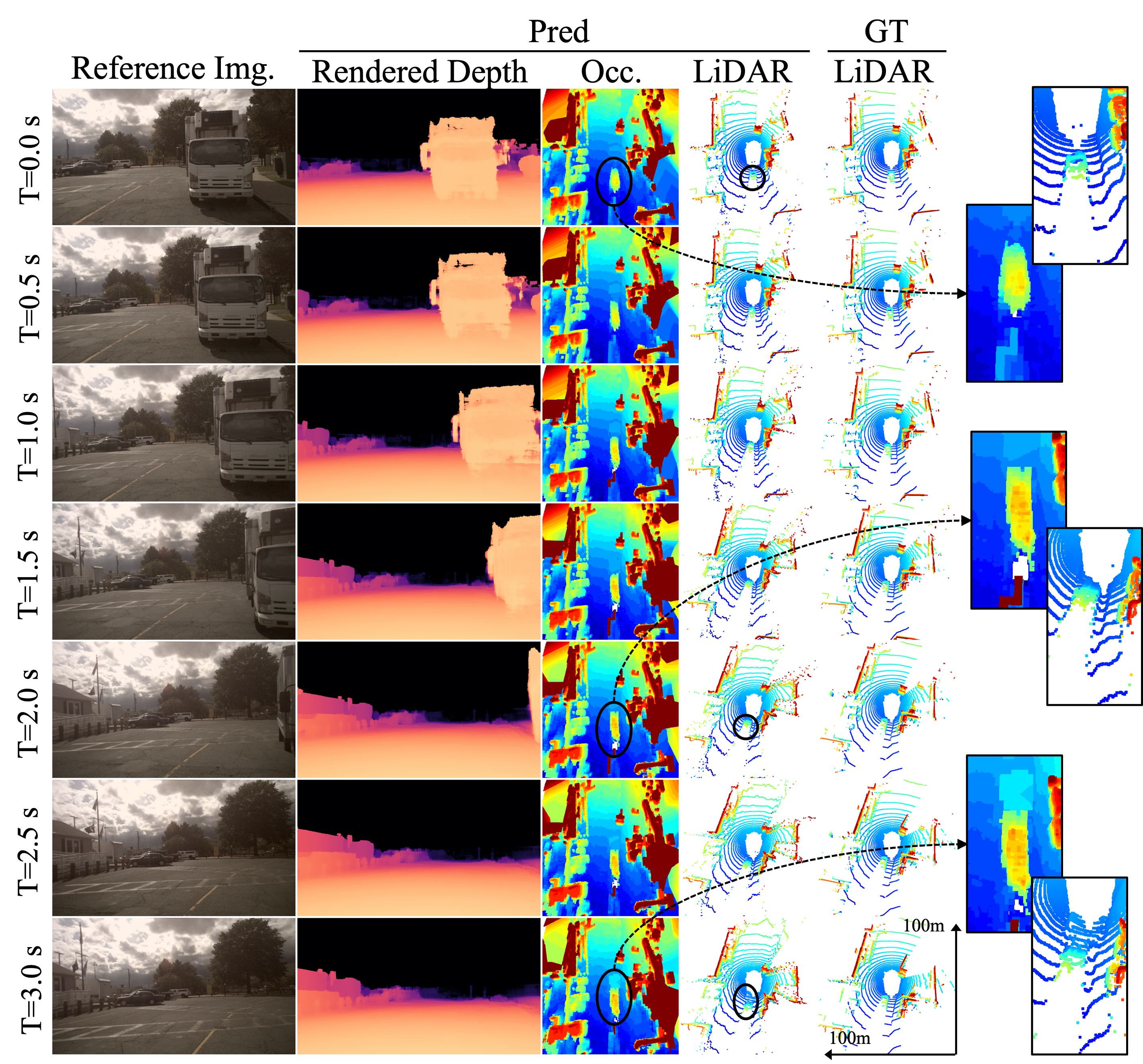}
     \caption{Future predictions up to 3 seconds of depth, occupancy, and LiDAR. The method captures fine details in the occupancy decoder, such as fine structures on the truck, forecasts the motion of other agents (evident in the trailing car), and expresses uncertainty through a gradual spread of occupancy. Please zoom in for details.} 
    \label{fig:QualitativeNuscenes}
    \vspace{-5mm}
\end{figure}

\subsection{Ablation Experiments}
\begin{table}[!tb]
\centering
\caption{Depth ablations. All methods use two iterative refinement steps. “DI” = depth-integration.}
\vspace{-0.5em}

\setlength{\tabcolsep}{3pt}
\renewcommand{\arraystretch}{0.9}

\resizebox{\linewidth}{!}{
\begin{tabular}{l|ccccc|cccc}
\toprule
\multirow{2}{*}{\textbf{Method}} &
\multirow{2}{*}{\textbf{Backbone}} &
\textbf{DINOv2} &
\textbf{Update} & \textbf{DI} & \textbf{Cam} &
\multirow{2}{*}{\textbf{MAE}$\downarrow$} &
\multirow{2}{*}{\textbf{RMSE}$\downarrow$} &
\multirow{2}{*}{\textbf{Time}$\downarrow$} &
\multirow{2}{*}{\textbf{MEM}$\downarrow$} \\
&& \textbf{dist.} & \textbf{Block} & \textbf{Module} & \textbf{Token} & &&& \\ 
\midrule
{OGNI-DC}               & CF \cite{zhang2023completionformer} & false & GRU & OGNI-DC & No  & 4.76 & 13.16 & 364ms & 2.4GB \\
\textbf{Ours}                 & Vim \cite{visionmamba}              & No & GRU & OGNI-DC & No  & 3.83 & 11.14 & 245ms & 1.3GB \\
\textbf{Ours}                              & Vim \cite{visionmamba}              & Yes  & GRU & OGNI-DC & No  & 3.61 & 10.32 & 245ms & 1.3GB \\
\textbf{Ours}                              & Vim \cite{visionmamba}              & Yes  & MGU & OGNI-DC & No  & 3.58 & 10.41 & 213ms & 1.3GB \\
\textbf{Ours}                              & Vim \cite{visionmamba}              & Yes  & MGU & Ours    & No  & 3.51 &  9.31 &  63ms & 1.3GB \\
\textbf{Ours}                              & ResNet50                             & Yes  & MGU & Ours    & No  & 3.53 &  9.28 & \textbf{58ms} & 2.9GB \\
\textbf{Ours}                              & Vim \cite{visionmamba}              & Yes  & MGU & Ours    & Yes & \textbf{3.46} & \textbf{9.21} & 63ms & \textbf{1.3GB} \\
\bottomrule
\end{tabular}}
\label{tab:depth_model_ablation}
\vspace{-0.5em}
\end{table}

\begin{table}[t]
    \centering
    \caption{Ablation Experiments. (a) Backbone Ablation, (b) History Horizon Ablation, (c) Self-Supervision Decoders.}
    \vspace*{-0.7em}
    \begin{minipage}[t]{0.48\linewidth}
        \centering
        \subcaption{Backbone Ablation}
        \label{tab:OD_backbone_ablation}
        \resizebox{\linewidth}{!}{%
            \begin{tabular}{lc|c}
            \toprule
                \textbf{Backbone} & \textbf{Pre-train} & \textbf{mAP} $\uparrow$ \\
                \midrule
                ResNet50     & stage 2 & 50.22 \\
                VisionMamba  & stage 2 & \textbf{52.61} \\
            \bottomrule
            \end{tabular}
        }
        \vspace*{0.01em}

        \subcaption{Self-Supervision Decoders}
        \label{tab:ablation_decoder}
        \resizebox{\linewidth}{!}{%
            \begin{tabular}{cc|cc}
            \toprule
                \textbf{\shortstack{Occupancy \\ Decoder}} & \textbf{\shortstack{Velocity \\ Decoder}} & \textbf{CD - 1s} $\downarrow$ & \textbf{CD - 3s} $\downarrow$ \\
                \midrule
                Yes & No  & 16.592 & 42.152 \\
                Yes & Yes & \textbf{15.821} & \textbf{39.031} \\
            \bottomrule
            \end{tabular}
        }
    \end{minipage}
    \hfill
    \begin{minipage}[t]{0.48\linewidth}
        \centering
        \subcaption{History Horizon Ablation}
        \label{tab:OD_past_ablation}
        \resizebox{\linewidth}{!}{%
            \begin{tabular}{c|c|c}
            \toprule
                \textbf{History H} & \textbf{Pre-train} & \textbf{mAP} $\uparrow$ \\
                \midrule
                0s & stage 1 & 50.75 \\
                \midrule
                \multirow{3}{*}{1s} & stage 1 & 49.58 \\
                                    & stage 2 & 52.61 \\
                                    & Improvement & \textbf{+6.11\%} \\
                \midrule
                \multirow{3}{*}{3s} & stage 1 & 51.16 \\
                                    & stage 2 & 51.86 \\
                                    & Improvement & \textbf{+1.37\%} \\
            \bottomrule
            \end{tabular}
        }
    \end{minipage}
    \label{tab:combined_ablations}
    \vspace{-1.0em}
\end{table}

We conduct a series of ablation studies to evaluate the design choices of the proposed architecture. 

\PAR{Depth Model.}
Table \ref{tab:depth_model_ablation} reports ablation experiments validating the proposed architectural components in terms of accuracy gains, inference time reduction, and memory footprint. Starting from \cite{ognidc}, replacing their backbone with our proposed Vim \cite{visionmamba} encoding leads to a notable reduction of memory of 45\% while boosting accuracy in MAE by 24\%. Further replacing the GRU update block with MGU improves inference time by 41\% over \cite{ognidc}, while integrating our proposed Depth Integration Module further reduces it to a total improvement of 82\%. Finally, adding Camera Tokens in the image encoding leads to a final improvement of 27\% in MAE over the baseline. Using the for BEV tasks commonly deployed ResNet50 instead of Vim decreases inference time marginally but doubles the memory footprint. As scaling to long ranges requires a lightweight backbone, we adopt the Vim-based model.

\PAR{Camera Backbone.}
Tab. \ref{tab:OD_backbone_ablation} reports mAP performance on the OD task with Vim \cite{visionmamba} backbone and with ResNet50 backbone. Our model with Vim improves mAP by $4.75\%$ over the proposed model with ResNet image backbone.

\PAR{History Horizon.}
In Tab. \ref{tab:OD_past_ablation} we ablate the input history horizon. Departing from existing work on urban scenarios \cite{li2022bevformer}, we empirically find that using a horizon of $1$ second on the long-range dataset leads to better Object Detection performance ($+1.44\%$) than $3$ seconds. This can be explained by the higher speeds involved in highway scenarios: higher velocities imply larger motions between frames during which instances can fall out of ROI and, in general, make the feature alignment harder. 
A  car at highway speeds travels in 3 seconds almost the entire region of interest of the NuScenes dataset \cite{caesar2020nuscenesmultimodaldatasetautonomous} ($\sim 100m$).

\PAR{Velocity Decoder.}
In Tab. \ref{tab:ablation_decoder}, we ablate the velocity decoder during pre-training. Training stage 2 with a 1-s input shows that velocity supervision improves LiDAR forecasting by 4.6\%, improving the encoding of moving objects, leading to more accurate predictions of dynamic actors.

\section{Conclusion}
We introduce a long-range camera-LiDAR BEV method that learns a sparse voxel representation for efficient and spatio-temporal 3D scene understanding.
To tackle the need for a large amount of training data, we devise a self-supervised pretraining approach that integrates temporal cues from past frames, enabling the model to predict future occupancy through supervision from raw sensor data. While existing BEV perception methods have been limited to $100m$, the method achieves long-range object detection up to $250m$, improving mAP by $\emph{26.6\%}$ ($\emph{+11.06 mAP}$) at distances up to $250m$. Our method also sets a new state-of-the-art in LiDAR forecasting, reducing Chamfer Distance by up to $\emph{30.5\%}$ across the full [$-100m:+250m$] range and outperforming existing work on the hardest 3-second future horizon in NuScenes by \emph{29\%}. We find that tackling highway scenarios for trucking is fundamentally different than urban perception, thus opening the avenue for a new line of work, specializing in achieving real-time long-range perception high-resolution sensors.
\vspace{-0.2em}

\section{Acknowledgments}
\vspace{-0.2em}
Felix Heide was supported by an NSF CAREER Award (2047359), a Packard Foundation Fellowship, a Sloan Research Fellowship, a Sony Young Faculty Award, a Project X Innovation Award, a Amazon Science Research Award, and a Bosch Research Award.

{
    \small
    \bibliographystyle{ieeenat_fullname}
    \bibliography{main}
}

\end{document}